\documentclass{article}
\usepackage{arxiv}

\usepackage[utf8]{inputenc} 
\usepackage[T1]{fontenc}    
\usepackage{palatino}
\usepackage{hyperref}       
\usepackage{url}            
\usepackage{booktabs}       
\usepackage{amsfonts}       
\usepackage{nicefrac}       
\usepackage{microtype}      
\usepackage{xcolor}         

\definecolor{darkblue}{rgb}{0, 0, 0.5}
\hypersetup{
    colorlinks=true,
    citecolor=darkblue,
    linkcolor=darkblue,
    urlcolor=darkblue,
}

\usepackage{amsmath,amsfonts,bm}

\def\eqref#1{equation~\ref{#1}}

\def\1{\bm{1}}

\DeclareMathAlphabet{\mathsfit}{\encodingdefault}{\sfdefault}{m}{sl}
\SetMathAlphabet{\mathsfit}{bold}{\encodingdefault}{\sfdefault}{bx}{n}

\usepackage[verbose=true,letterpaper]{geometry}
\usepackage{natbib}
\usepackage[hyphenbreaks]{breakurl} 
\usepackage{float}
\usepackage{xspace}
\usepackage{enumerate}
\usepackage{graphicx}
\usepackage{pgf}
\usepackage{paralist}
\usepackage{wrapfig}
\usepackage{nicefrac}       

\setcitestyle{authoryear,open={(},close={)}}

\usepackage{fullpage}

\usepackage{graphicx}
\usepackage{bbm}
\usepackage{enumitem}
\usepackage{multirow}
\usepackage{setspace}
\usepackage{cleveref}

\usepackage{pifont}         
\NewDocumentCommand{\drsh}{ O{0.6em} O{0.5em} O{0.65pt} O{rounded corners=1.6pt} }{%
  \mathrel{%
  \hspace{0.3em}
    \tikz[baseline=-0.6em]{%
      \draw[->, line width=#3, #4] (0,0) -- (0,-#2) -- (#1,-#2);
    }%
  }%
}

\usepackage{amsmath}
\usepackage{amssymb}
\usepackage{mathtools}
\usepackage{amsthm}

\theoremstyle{plain}

\theoremstyle{remark}

\usepackage{listings}
\usepackage{siunitx}
\usepackage{bookmark}
\usepackage{textcomp}
\usepackage{nth}
\usepackage{indentfirst}
\usepackage{siunitx}
\usepackage{changepage}
\usepackage{bm}
\usepackage{multirow}
\usepackage{multicol}
\usepackage{colortbl}

\usepackage{subfig}
\usepackage{makecell}

\usepackage{titling}
\usepackage{setspace}

\usepackage[most,skins,theorems,breakable]{tcolorbox}
\tcbset{
  mybox/.style={
    width=\linewidth,
    top=14pt,
    bottom=7pt,
    colback=cyan!5!white,
    colframe=black,
    colbacktitle=black,
    enhanced,
    center,
    breakable,
    attach boxed title to top left={yshift=-0.13in,xshift=0.15in},
    boxed title style={boxrule=0pt,colframe=white,},
  }
}
\newtcolorbox{mybox}[2][]{mybox,title=#2,#1}

\usepackage{algorithm}
\usepackage{algorithmicx}
\usepackage{algpseudocode}

\newcommand{\ourmethod}{STT\xspace}
\newcommand{\ourbench}{STT-Bench\xspace}
\newcommand{\STT}{Sleep-Time Training\xspace}

\newcommand{\QFourB}{Qwen3.5-4B\xspace}
\newcommand{\QNineB}{Qwen3.5-9B\xspace}
\newcommand{\QThirtyB}{Qwen3-30B-A3B-Instruct-2507\xspace}

\colorlet{softred}{red!80!black}
\colorlet{softgreen}{green!55!black}
\colorlet{softblue}{blue!65!black}
\colorlet{softyellow}{yellow!70!black}
\colorlet{softgray}{black!50}
\definecolor{softblue}{RGB}{38, 139, 210}

\newcommand{\cmark}{\textcolor{softgreen}{\ding{51}}}
\newcommand{\xmark}{\textcolor{softred}{\ding{55}}}
\newcolumntype{M}[1]{>{\centering\arraybackslash}m{#1}}

\usepackage{array}
\newcolumntype{L}[1]{>{\raggedright\arraybackslash}p{#1}}
\newcolumntype{C}[1]{>{\centering\arraybackslash}p{#1}}

\newcommand\blfootnote[1]{%
  \begingroup
  \renewcommand\thefootnote{}\footnote{#1}%
  \addtocounter{footnote}{-1}%
  \endgroup
}

\title{\textbf{Sleep-Time Training: From Deployment Experience to Persistent Agent Improvement}}
\author{
{{\textbf{Runze Liu$^{1}$ ~ Haoran He$^{1}$ ~ Jiashun Liu$^{1}$ ~ Zijie Geng$^{2}$ ~ Yuqian Fu$^{3}$ ~ Wang You$^{4}$}}}\\
{{\textbf{Chen Hu$^{4}$ ~ Daxin Jiang$^{4}$ ~ Ling Pan$^{1\dag}$}}}\\
{\normalsize{$^{1}$Hong Kong University of Science and Technology}}\\
{\normalsize{$^{2}$State Key Laboratory of Multimodal Artificial Intelligence Systems, CASIA}}\\
{\normalsize{$^{3}$University of Science and Technology of China $^{4}$StepFun}}\\
{\normalsize{\nolinkurl{runze.liu@connect.ust.hk} \quad \nolinkurl{lingpan@ust.hk}}}
}
\date{}

\begin{document}

\maketitle

\vspace{-3.0em}
\blfootnote{$^\dag$ Corresponding author: Ling Pan}

\begin{abstract}
Foundation agents are increasingly deployed as persistent personal assistants that serve the same user across many sessions, yet their policies typically remain frozen after deployment. Corrections, latent preferences, and effective strategies discovered through interaction remain largely external to the policy: at best logged externally and retrieved reactively or rely on extra inference-time computation, rather than internalized to proactively shape future behavior. We introduce Sleep-Time Training (STT), a modality-agnostic, proactive deployment-time learning paradigm that consolidates daily interaction experience into persistent policy updates before the user returns. However, deployment trajectories collected during active sessions are noisy and rarely successful, leaving standard outcome-based RL with almost no effective learning signal. STT overcomes this critical challenge in two stages: (i) an auxiliary model reconstructs environment-verified demonstrations for supervised fine-tuning, recalibrating action priors toward viable execution paths; (ii) iterative step repair localizes the earliest defective intermediate steps to construct original--repaired trajectory pairs, supplying dense, off-policy contrast for group-relative reinforcement learning. To evaluate continual adaptation across repeated sessions in a controlled stateful testbed, we build STT-Bench, a four-day benchmark comprising 20 multi-day base task families and 20 held-out generalization families under varying levels of preference disclosure. Across three model scales and two architectures, STT significantly outperforms baselines without increasing online inference cost. These gains persist across consecutive sleep cycles and transfer to unseen workflows, opening a new avenue for continual agent learning from deployment experience.
\end{abstract}

\section{Introduction}
\label{sec:introduction}

Foundation agents across language, vision-language, and embodied domains are increasingly deployed as persistent personal assistants that navigate digital interfaces, invoke tools, edit code, and manage multi-step workflows~\citep{GPT-5.6, Claude-Fable-5, Qwen3.5}.
Personal agents such as OpenClaw~\citep{OpenClaw}, Codex~\citep{Codex}, and Claude Code~\citep{Claude-Code} demonstrate that modern agents can execute increasingly complex workflows.
In the natural deployment setting of personal agents, a user may ask an agent to organize files, debug a codebase, book travel, or maintain a project; the agent acts through tools, receives corrections, sometimes fails, and the user leaves. Hours or days later, the user returns with related goals, unfinished workflows, and recurring latent preferences.
This repeated-session lifecycle makes deployment itself a rich source of experience for learning~\citep{silver2025welcome}.

Despite this rich stream of experience, agents are largely static between different sessions in current deployment paradigms~\citep{OpenClaw}. Their parameters are fixed after large-scale pre-training and post-training, even as they observe user corrections, implicit preferences, tool failures, and successful strategies. As a result, an agent may spend one session discovering how a user wants reports formatted or which coding style a project follows, only to require similar guidance again in the next session, or passively log them into external memory without internalizing them into the policy parameters. Consequently, personal agents operate in evolving, user-specific workflows while their underlying policies remain frozen.

A natural question therefore arises: \emph{how should personal agents continue to improve during deployment?} Existing approaches remain fundamentally \emph{reactive}, operating along two axes. First, test-time scaling reacts to individual queries on the fly by allocating extra inference compute, such as sampling multiple reasoning paths and selecting the best one~\citep{snell2025scaling, liu2025can}. While this can improve the immediate response, it increases user-facing latency and yields only transient gains. Second, external memory methods rely on \emph{reactive memory}: they log facts, trajectories, or distilled strategies in external stores and retrieve them when a similar situation arises~\citep{Mem0, ReasoningBank, MemEvolve, ACE}. Even when agents offline reorganize transcripts or refine notes between sessions, the learned knowledge stays outside the model parameters. Consequently, adaptation remains bottlenecked by retrieval accuracy, prompt context limits, and explicit user cues, much like consulting an external notebook rather than mastering a skill.

In human cognition, rapid online encoding of daily experiences is complemented by offline consolidation during sleep, where episodic traces are replayed and integrated into long-term neocortical knowledge. Inspired by this division of labor, we advocate shifting agent adaptation from \emph{reactive memory} to \emph{proactive learning}. Rather than passively indexing past logs for future lookup, an agent should use naturally occurring idle periods between sessions (``sleep-time'', such as overnight) to reflect backward on recent interactions, repair mistakes, and consolidate deployment experience directly into its policy weights before the user returns. Under this paradigm, \emph{Sleep-Time Training} (\ourmethod) turns the repeated-session lifecycle of personal agents into a continual weight-level learning loop: the agent collects real trajectories during active ``day-time'' use and proactively updates its internalized policy during sleep-time so it can act competently and follow user preferences forward without test-time retrieval overhead.

Real deployment trajectories, however, cannot be naively repurposed for parameter updates. They often suffer from cascading tool-use failures, redundant actions, and sparse outcome signals. Moreover, when the current policy rarely completes a complex workflow successfully, standard outcome-based RL~\citep{GRPO, DAPO, Archer, ASPO} collapses due to a lack of positive learning signals. The key challenge is therefore how to transform sparse, imperfect traces into reliable and sufficiently dense supervision for persistent policy improvement.

To resolve this challenge, we develop a two-stage sleep-time training procedure. First, an auxiliary model reconstructs imperfect deployment trajectories into corrected demonstrations, which are replayed and verified in the environment before being incorporated into supervised fine-tuning. This stage rapidly recalibrates the agent's action priors toward viable execution paths. Second, we perform iterative step repair for reinforcement learning. \ourmethod locates the earliest defective assistant action and replaces only that single step, allowing the agent to roll out the remainder of the trajectory autonomously. We then perform group-relative RL over pairs of original trajectories and repaired counterparts, providing dense supervision while keeping the agent's own interaction dynamics in the loop. Unlike test-time scaling, which shifts computational overhead onto waiting users, \ourmethod strengthens the agent offline before the next query arrives. Unlike context-level sleep-time compute~\citep{Sleep-Time-Compute}, which pre-computes anticipated answers or reorganizes external notes for reactive retrieval, \ourmethod internalizes both general execution reliability and latent user preferences directly into the policy. Table~\ref{tab:deployment_time_paradigms} summarizes these distinctions across user latency, experience usage, policy updates, and persistent gains.

Evaluating this setting requires benchmarks beyond single-session task success. A sleep-time training benchmark should test whether an agent improves after prior interactions, retains gains across days, infers hidden user preferences, and transfers experience to related unseen tasks. To isolate multi-day policy consolidation from low-level perceptual noise, we introduce \ourbench, a controlled stateful benchmark designed to measure cross-day retention, preference internalization under progressively reduced prompt cues, and transfer to unseen workflows. Across multiple models, \ourmethod consistently outperforms memory-based and RL baselines. The gains persist across sleep cycles, transfer to held-out tasks, and remain substantial when the prompt no longer states the user's preferences, demonstrating that deployment trajectories can be converted into reusable learning signals for continual agent improvement.

Our contributions are as follows:
\begin{itemize}[leftmargin=*, topsep=0pt, parsep=6pt, partopsep=0pt]
\item We introduce \textit{\STT} (STT), a novel deployment-time learning paradigm that converts interaction experience into parameter updates during user-idle periods, enabling persistent agent improvement without increasing user-facing inference cost.
\item We design a two-stage sleep-time training framework that turns noisy deployment traces into dense learning signals by recalibrating action priors via environment-replayed demonstrations before refining recovery behavior through localized single-step repair and group-relative reinforcement learning.
\item To evaluate the sleep-time training capabilities of personal agents, we introduce \ourbench, a four-day benchmark with 20 multi-day base task families, 20 held-out generalization families, and controlled levels of preference disclosure. Experiments show that STT outperforms memory-based and RL baselines, retains improvements across days, and transfers to both future task instances and unseen task families.
\end{itemize}

\begin{table*}[!t]
\centering
\caption{Comparison of deployment-time improvement paradigms.
\ourmethod uniquely turns deployment experience into persistent policy improvement
during user-idle time.}
\label{tab:deployment_time_paradigms}
\resizebox{\textwidth}{!}{
\begin{tabular}{L{7.2cm}cccc}
\toprule
\textbf{Paradigm}
& \textbf{User Latency}
& \textbf{Uses Deployment Experience}
& \textbf{Updates Policy}
& \textbf{Persistent Gain} \\
\midrule
\makecell[l]{Test-Time Scaling\\[-1pt]
\citep{Self-Consistency,snell2025scaling}}
& \xmark & \xmark & \xmark & \xmark \\
\makecell[l]{External Memory / Skills\\[-1pt]
\citep{Mem0,ReasoningBank}}
& \cmark & \cmark & \xmark & \xmark \\
\makecell[l]{Sleep-Time Scaling\\[-1pt]
\citep{Sleep-Time-Compute}}
& \cmark & \xmark & \xmark & \xmark \\
\midrule
\textbf{Sleep-Time Training (Ours)}
& \cmark & \cmark & \cmark & \cmark \\
\bottomrule
\end{tabular}
}
\vspace{-.2in}
\end{table*}

\section{Related Work}
\label{sec:related_work}

\paragraph{Evolution of LLM Agents.}

Reinforcement learning (RL) has become a central paradigm for improving the multi-step reasoning and tool-use capabilities of LLM agents~\citep{zhang2025survey, ARPO}. Recent work explores online and periodic policy optimization for personal agents: OpenClaw-RL~\citep{OpenClaw-RL} trains agents using process rewards and feedback-conditioned distillation from subsequent interaction states, MetaClaw~\citep{MetaClaw} co-evolves a reusable skill library alongside RL updates during user-inactive windows, and Claw-R1~\citep{Claw-R1} provides step-level data infrastructure for curating agent rollouts. Complementary efforts improve agents from the data and environment side by evolving shared skill libraries across users~\citep{SkillClaw} or synthesizing training environments and SFT/RL trajectories~\citep{Agent-World, ClawGym}. Closely related to our trajectory repair mechanism, Agent-R~\citep{Agent-R} uses Monte Carlo tree search to construct revision trajectories for iterative self-training, while SDPO~\citep{SDPO} distills feedback-conditioned predictions into the policy. In contrast, \ourmethod combines verified demonstration reconstruction with group-relative RL over original trajectories and agent continuations following localized step repair.

\paragraph{Deployment-Time Improvement.}

Existing deployment-time methods fall primarily into test-time adaptation~\citep{snell2025scaling, liu2025can, TTRL} and context-based sleep-time processing~\citep{Sleep-Time-Compute}. Test-time scaling allocates extra inference compute per query, such as via self-consistency~\citep{Self-Consistency} or compute-optimal search guided by process reward models~\citep{snell2025scaling, liu2025can}. GenPRM~\citep{GenPRM} further scales process verification by performing explicit reasoning and code verification before judging each step. Test-time training instead updates parameters during inference using retrieved neighbors~\citep{TTT-NN} or majority-vote rewards~\citep{TTRL}. However, these approaches either increase user-facing latency or rely on transient per-query updates. Conversely, sleep-time compute~\citep{Sleep-Time-Compute}, ProAct~\citep{ProAct}, and offline log-reflection methods pre-process context or reorganize external memory during idle periods, keeping adaptation reactive at inference time and leaving the underlying policy parameters unchanged. \ourmethod bridges this gap by shifting from reactive context organization to proactive policy consolidation, using idle periods to internalize deployment experience directly into model weights via verified trajectory repair and reinforcement learning.

\section{STT-Bench}

Existing tool-use benchmarks primarily measure task completion within a single interaction~\citep{PinchBench, Claw-Eval, WildClawBench}, which is insufficient for evaluating our new sleep-time training paradigm: an agent may complete a task by following explicit instructions, yet fail to internalize user-specific preferences, transfer them across days, or improve when the next prompt omits those preferences.
\ourbench is designed to evaluate this missing capability.

Specifically, \ourbench is built around three design principles.
First, tasks span multiple chronological days to measure whether deployment experience leads to persistent improvement in later interactions.
Second, we vary how explicitly user preferences are revealed in the current prompt to test whether the agent has learned these preferences from previous interactions.
Third, we include held-out generalization task families to evaluate whether learned user-specific behavior transfers to related but unseen workflows.

\paragraph{Persona and Task Synthesis.}

We construct \ourbench from realistic tool-use scenarios. We first collect documentation for 256 frequently used Model Context Protocol (MCP) servers from Smithery\footnote{\url{https://smithery.ai/}} and retain their server descriptions and tool definitions after data cleaning.

We use Codex to synthesize user personas, task families, mock services, and executable task instances. The persona describes recurring user preferences that can affect how a task should be completed, such as preferred communication style, formatting conventions, workflow choices, or constraints on external actions.

To separate memorization from transfer, \ourbench contains 40 task families, split into 20 \emph{base} families and 20 \emph{generalization} families. Base families are available during sleep-time learning and evaluation. Generalization families are used only for evaluation. Each generalization family shares user-level habits or skills with the base split but requires a different task scenario or workflow, allowing us to test whether learned behavior transfers beyond previously observed tasks.

\begin{wraptable}[8]{r}{0.54\columnwidth}
\vspace{-0.4em}
\centering
\footnotesize
\caption{Preference-disclosure levels in \ourbench.}
\begin{tabular}{@{}cL{0.39\columnwidth}@{}}
\toprule
\textbf{Level} & \textbf{Prompt disclosure} \\
\midrule
\textbf{L0} & All preferences are explicit. \\
\textbf{L1} & One preference is explicit. \\
\textbf{L2} & Prior habits are referenced indirectly. \\
\textbf{L3} & No preference hint is provided. \\
\bottomrule
\end{tabular}
\label{tab:prompt_levels}
\end{wraptable}
Each base family contains four chronological instances, one for each day, while each generalization family contains one held-out Day~4 instance, yielding 100 task instances in total. Every instance uses the same user persona. We additionally construct four prompt levels, summarized in Table~\ref{tab:prompt_levels}, that vary how explicitly the prompt reveals the user's preferences.
Higher levels therefore place more emphasis on learning user-specific behavior from earlier interactions rather than from the current prompt.

\paragraph{Evaluation Protocol.}
\label{sec:metrics}

To enable reproducible and safe evaluation, all tasks are executed against mock services rather than live external services, following the general design of Claw-Eval~\citep{Claw-Eval}. The agent can interact with the environment only through the declared tool interfaces.

Rather than judging final text responses alone, we compute a unified score from service-side audit logs and final task artifacts:
\begin{equation}
    \label{eq:benchmark_score}
    \operatorname{Score}(\tau) = C(\tau) \cdot E(\tau) \cdot P(\tau),
\end{equation}
where $C(\tau) \in [0, 1]$ measures functional task completion (weighted checks on entities, state mutations, and artifact quality), $E(\tau) = \min(1, T_{\mathrm{ref}}/T) \in [0, 1]$ penalizes redundant assistant steps relative to the reference count $T_{\mathrm{ref}}$, and $P(\tau) \in [0, 1]$ averages adherence across the three user preferences. We report this multiplicative composite score throughout our experiments so that high performance requires completing the workflow efficiently while respecting user preferences, and we isolate preference retention by progressively removing prompt hints (L1--L3) in Section~\ref{sec:analysis}. Appendix~\ref{app:eval_details} provides the detailed scoring rules.

\paragraph{Task Validation.}

Task generation is followed by automatic execution and human verification. During synthesis, the coding agent executes the reference workflow against the mock service and checks whether the task, tools, and verifier are mutually consistent. Invalid tool schemas, unreachable states, ambiguous requirements, and inconsistent verifier conditions are repaired before the task is admitted to the benchmark.

We then perform two sanity checks on the evaluator. First, the reference trajectory must obtain an overall score equal to $1.0$. Second, a null trajectory that performs no meaningful task action must obtain a score below $0.1$. These checks detect overly permissive or incorrectly specified evaluators.

Finally, we manually inspect all benchmark families to verify that: (1) tasks are executable; (2) prompts do not leak hidden evaluation criteria; (3) evaluators align with the intended task objective; (4) user preferences are meaningful and consistently represented across days; and (5) base and generalization families do not share identical workflows. After filtering, \ourbench comprises 40 human-verified task families. Appendix~\ref{app:benchmark_details} details the construction and calibration pipeline.

\section{Sleep-Time Training}
\label{sec:method}

\subsection{Problem Setting}
\label{sec:setting}

Consider an agent with policy $\pi_{\theta_d}$ at the beginning of day $d$. While our formulation is agnostic to input modality (handling textual, visual, or structured environment feedback), we instantiate each step through tool-mediated interaction: given a task $x$, the agent observes the interaction history $h_t$ at step $t$, generates an action $a_t$, and receives an environment observation $o_t$. A complete trajectory is
\begin{equation}
    \tau=(x,a_1,o_1,\ldots,a_{T-1},o_{T-1},a_T),
\end{equation}
where $a_T$ contains the final user-facing response.

Let $\mathcal{H}_d$ denote the interaction trajectories collected on day $d$. During sleep time, the agent updates its parameters via
\begin{equation}
    \theta_{d+1}=U(\theta_d,\mathcal{H}_d;B_d),
    \label{eq:stt_update}
\end{equation}
where $B_d$ is the available sleep-time computation budget. Let $\mathcal{D}_{\mathrm{future}}$ denote the distribution of future user tasks and $S(x,\tau)$ the evaluation score of trajectory $\tau$ on task $x$. \ourmethod seeks an update rule $U$ that maximizes expected future performance:
\begin{equation}
    J_{\mathrm{STT}}(\theta_{d+1})
    =\mathbb{E}_{\substack{x\sim\mathcal{D}_{\mathrm{future}}\\
    \tau\sim\pi_{\theta_{d+1}}(\cdot\mid x)}}
    \left[S(x,\tau)\right],
    \label{eq:stt_objective}
\end{equation}
while keeping the online inference procedure and budget unchanged. Future tasks define this objective but are never observed during training. Instead, each daily update operates solely on current-day trajectories, carrying prior experience forward through the updated weights $\theta_d$ without replaying historical transcripts or expanding future prompts.

\begin{figure*}[!t]
\centering
\includegraphics[width=1.0\textwidth]{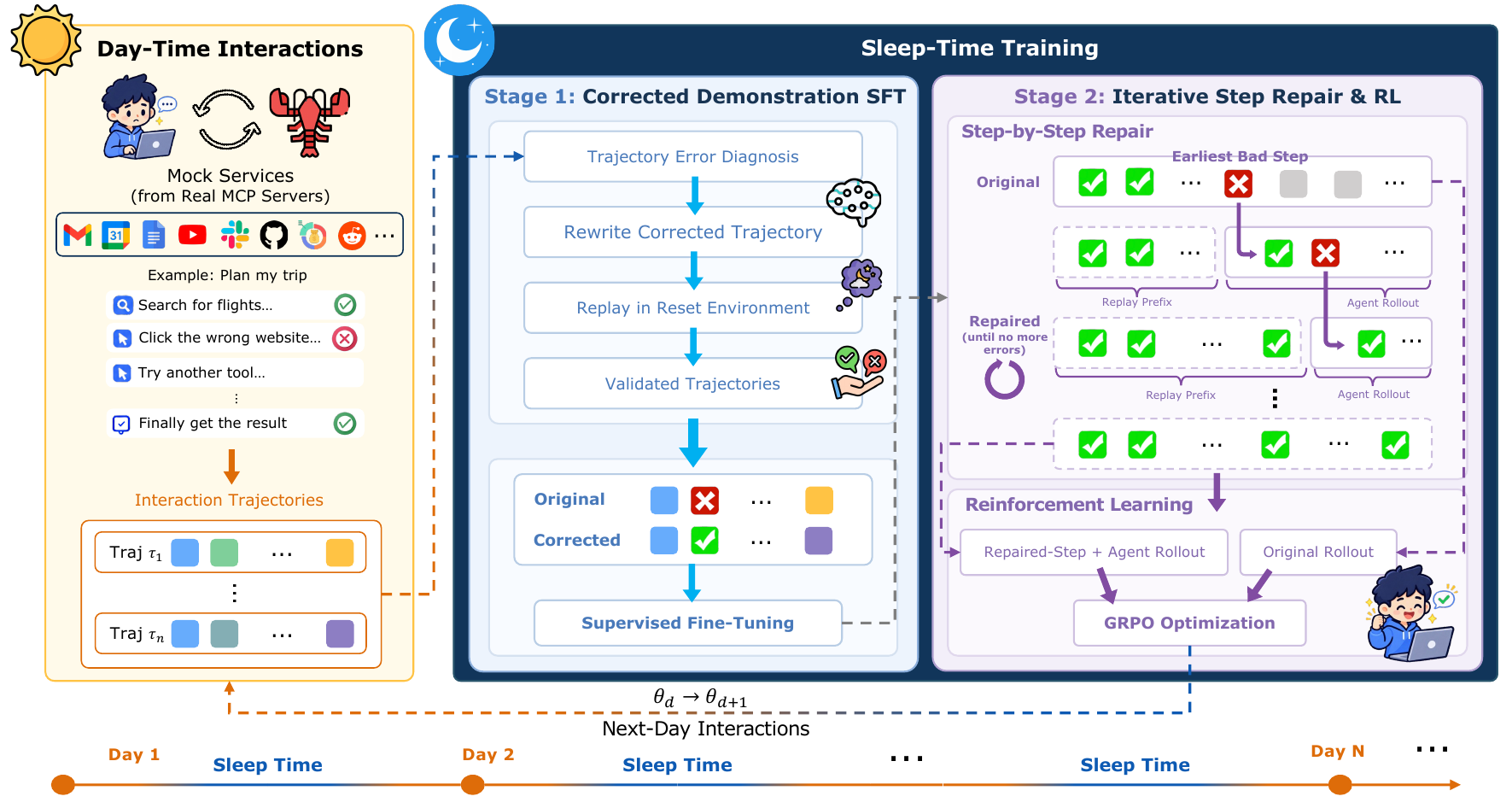}
\caption{Overview of \ourmethod. The agent collects interaction trajectories during user-facing sessions and converts them into policy updates during sleep time. First, an auxiliary model reconstructs, replays, and validates corrected demonstrations for SFT. Second, iterative step repair pairs original trajectories with repaired continuations for group-relative RL.}
\label{fig:overview}
\end{figure*}

\subsection{Learning from Corrected Demonstrations}
\label{sec:sft}

\paragraph{Reference collection.}
For each current-day task, we collect $N$ independent trajectories using $\pi_{\theta_d}$ from identical initial states, and select the first $M\leq N$ trajectories by sample index without consulting benchmark scores. Each selected trajectory serves as the reference trace for synthesizing one demonstration. These rollouts constitute the offline experience-collection budget for sleep-time updates.

\paragraph{Synthesis and replay.}
Given the public task specification, tool schemas, and a reference trajectory, an auxiliary model reconstructs a complete sequence of assistant actions that preserves valid steps while correcting errors. The auxiliary model never fabricates tool outputs: the environment is reset and the proposed actions are executed to obtain genuine observations. If execution or verification reveals an error, the candidate is iteratively revised and replayed, preserving valid prefixes before model-diagnosed errors or resynthesizing from rule feedback when deterministic environment checks fail.

A candidate demonstration is admitted into $\mathcal{D}_{\mathrm{SFT}}$ only if it passes public-task review, executable environment assertions derived solely from the visible prompt (without access to hidden preference rubrics or benchmark graders), and structural validity checks (grounded final response, valid tool schemas, zero unresolved execution errors, and adherence to step and context limits).

\paragraph{Supervised update.}
Representing each accepted demonstration as a context--target pair $(c,y)$, we optimize the standard supervised fine-tuning objective:
\begin{equation}
    \mathcal{L}_{\mathrm{SFT}}(\theta)
    =-\mathbb{E}_{(c,y)\sim\mathcal{D}_{\mathrm{SFT}}}
      \left[\log\pi_\theta(y\mid c)\right],
    \label{eq:sft}
\end{equation}
where the loss is computed strictly over assistant tokens, masking user prompts and tool observations. The resulting policy $\pi_{\mathrm{SFT}}$ initializes the subsequent RL stage.

\subsection{Sleep-Time RL via Iterative Step Repair}
\label{sec:repair}

For each RL update, we freeze the current policy as $\pi_{\mathrm{old}}$ and sample $K$ trajectories per task. The auxiliary model inspects each trajectory chronologically to locate the earliest defective assistant action at step $t^\star$ (such as an erroneous tool call, unsupported claim, omitted deliverable, or redundant loop) and proposes a single replacement action $\widetilde a_{t^\star}$ conditioned strictly on information available prior to step $t^\star$.

We then reset the environment, replay the prefix $a_{<t^\star}$, execute $\widetilde a_{t^\star}$, and allow the agent $\pi_{\mathrm{old}}$ to roll out the remainder of the trajectory autonomously:
\begin{equation}
    \widetilde\tau
    =\operatorname{Replay}(a_{<t^\star})
       \oplus\widetilde a_{t^\star}
       \oplus\operatorname{Continue}_{\pi_{\mathrm{old}}}(\widetilde h_{t^\star+1}),
    \label{eq:repair}
\end{equation}
where $\oplus$ denotes concatenation and all intermediate observations are produced by environment execution. The auxiliary model re-evaluates the resulting trajectory and repeats this localized repair loop until the trajectory passes review or reaches the step budget. Each completed repair yields a final pair $(\tau, \widetilde\tau)$ consisting of the original rollout and the last accepted revision, while intermediate revisions are discarded. Already valid trajectories remain paired with an identical copy, whereas pairs that fail to produce a valid revision within budget are excluded from the training group.

\paragraph{Verifiable outcome scoring.}
An independent scoring call evaluates each retained trajectory against the public task requirements and its observed execution trace, assigning a binary reward $R_i\in\{0,1\}$. Because the scorer evaluates each trajectory solely against the public task specification without seeing repair histories, hidden user preferences, or ground-truth evaluation rubrics, repaired trajectories earn positive rewards only when they genuinely resolve the task, while original trajectories that recover from early missteps on their own can still receive positive rewards.

\paragraph{Group Relative Policy Optimization (GRPO).}
For each task, the $0<G\leq 2K$ retained trajectories form a group $\{\tau_i\}_{i=1}^{G}$. Following GRPO~\citep{GRPO}, we compute group-normalized advantages $\widehat{A}_i=(R_i-\bar R)/\sigma_R$, skipping groups with zero reward variance. Let $y_{i,j}$ and $h_{i,j}$ denote the $j$-th assistant token and its preceding history, $L_i$ the assistant token length of $\tau_i$, and $r_{i,j}(\theta)=\pi_\theta(y_{i,j}\mid h_{i,j})/\pi_{\mathrm{old}}(y_{i,j}\mid h_{i,j})$. We minimize the clipped surrogate objective:
\begin{equation}
\mathcal{L}_{\mathrm{RL}}(\theta)
=-\frac{1}{G}\sum_{i=1}^{G}\frac{1}{L_i}\sum_{j=1}^{L_i}
\min\!\bigg(
r_{i,j}(\theta)\widehat{A}_i,\,
\operatorname{clip}\!\left(r_{i,j}(\theta),
1-\varepsilon_{\mathrm{low}},1+\varepsilon_{\mathrm{high}}\right)
\widehat{A}_i
\bigg),
\label{eq:rl}
\end{equation}
where token-level averaging within each trajectory prevents length bias. Because repaired trajectories combine auxiliary-model interventions with agent-generated continuations, optimizing~\eqref{eq:rl} acts as an off-policy, group-relative update that aligns the policy toward verified recovery paths while staying anchored to $\pi_{\mathrm{old}}$ via importance clipping.

\paragraph{Why combine the two stages?}
The two stages address complementary failure regimes. Early in deployment, weak initial policies rarely complete complex workflows, leaving outcome-based RL starved of positive reward signals; SFT on replayed demonstrations rapidly establishes viable execution priors. Once the policy is competent, iterative step repair exposes the agent to states induced by its own rollouts and provides fine-grained contrasts between failed and repaired trajectories. Requiring the agent to autonomously complete the suffix after each single-step edit prevents sleep-time learning from reducing to passive behavioral cloning of auxiliary trajectories. The complete multi-day procedure is summarized in Algorithm~\ref{alg:ste} in Appendix~\ref{app:full_algorithm}.

\section{Experiments}
\label{sec:exp}

\subsection{Setup}
\label{sec:exp_setup}

\begin{table*}[!t]
\centering
\caption{Evaluation results on \ourbench. The results of \ourmethod are shaded and the highest values are \textbf{bolded}.}
\resizebox{0.93\textwidth}{!}{
\begin{tabular}{lcccccccc}
\toprule
\multirow{2}{*}{\textbf{Method}} & \multicolumn{4}{c}{\bf Seen Tasks} & \multicolumn{3}{c}{\bf Unseen Tasks} & \multirow{2}{*}{\textbf{Avg.}} \\
\cmidrule(lr){2-5}\cmidrule(lr){6-8}
& {\footnotesize\color{gray}{Day 1}} & {\footnotesize\color{gray}{Day 2}} & {\footnotesize\color{gray}{Day 3}} & {\footnotesize\color{gray}{Avg.}} & {\footnotesize\color{gray}{Day 4 Base}} & {\footnotesize\color{gray}{Day 4 Gen.}} & {\footnotesize\color{gray}{Avg.}} & \\
\midrule
\textbf{\QFourB} &  &  &  &  &  &  &  &  \\
$\drsh$ ReasoningBank & 24.4 & 22.3 & 28.3 & 25.0 & 22.8 & 10.8 & 16.8 & 20.9 \\
$\drsh$ MemEvolve     & 25.0 & 25.0 & 25.4 & 25.1 & 20.9 & 11.1 & 16.0 & 20.6 \\
$\drsh$ OpenClaw-RL   & 25.9 & 25.4 & 25.3 & 25.5 & 23.5 & 12.8 & 18.2 & 21.8 \\
\rowcolor{cyan!10} $\drsh$ \ourmethod (Ours) & \textbf{45.4} & \textbf{48.4} & \textbf{50.6} & \textbf{48.1} & \textbf{40.3} & \textbf{26.4} & \textbf{33.4} & \textbf{40.8} \\
\midrule
\textbf{\QNineB} &  &  &  &  &  &  &  &  \\
$\drsh$ ReasoningBank & 33.5 & 33.1 & 33.4 & 33.3 & 34.2 & 19.1 & 26.6 & 30.0 \\
$\drsh$ MemEvolve     & 35.0 & 33.8 & 31.6 & 33.4 & 34.1 & 19.4 & 26.8 & 30.1 \\
$\drsh$ OpenClaw-RL   & 36.2 & 37.2 & 32.6 & 35.3 & 34.8 & 21.0 & 27.9 & 31.6 \\
\rowcolor{cyan!10} $\drsh$ \ourmethod (Ours) & \textbf{46.0} & \textbf{47.2} & \textbf{50.8} & \textbf{48.0} & \textbf{41.5} & \textbf{26.6} & \textbf{34.0} & \textbf{41.0} \\
\midrule
\textbf{\QThirtyB} &  &  &  &  &  &  &  &  \\
$\drsh$ ReasoningBank & 14.4 & 13.9 & 14.8 & 14.4 & 15.0 & 12.2 & 13.6 & 14.0 \\
$\drsh$ MemEvolve     & 17.4 & 13.2 & 12.8 & 14.5 & 14.4 & 11.7 & 13.1 & 13.8 \\
$\drsh$ OpenClaw-RL   & 13.8 & 15.2 & 16.4 & 15.1 & 13.2 & 11.2 & 12.2 & 13.7 \\
\rowcolor{cyan!10} $\drsh$ \ourmethod (Ours) & \textbf{24.1} & \textbf{21.1} & \textbf{24.0} & \textbf{23.1} & \textbf{22.2} & \textbf{15.4} & \textbf{18.8} & \textbf{21.0} \\
\bottomrule
\end{tabular}%
}
\label{tab:main}%
\end{table*}%

\paragraph{Models and Baselines.}

We conduct experiments using Qwen3.5-4B, Qwen3.5-9B, and Qwen3-30B-A3B-Instruct-2507 as the base models~\citep{Qwen3, Qwen3.5}. We use non-thinking mode for all three models.
We compare \ourmethod against several representative baselines.
All baselines follow the same three-day schedule.
\begin{itemize}[leftmargin=*]
    \item \textbf{ReasoningBank}~\citep{ReasoningBank}: Retrieves strategies from successful and failed trajectories.
    \item \textbf{MemEvolve}~\citep{MemEvolve}: Updates memory content and its management code. We use one candidate per day, with 16 base families for memory construction and 4 for validation using public-task success.
    \item \textbf{OpenClaw-RL}~\citep{OpenClaw-RL}: We use Binary RL with four trajectories per task and four steps per day. Step rewards use majority voting over three judgments from tool feedback.
\end{itemize}
We use GPT-5.6-Luna for all auxiliary-model calls, including memory operations for ReasoningBank and MemEvolve, process-reward inference for OpenClaw-RL, and benchmark judging for all three methods.

\paragraph{Training Details.}

We implement \ourmethod with slime v0.3.0~\citep{slime} and use SGLang~\citep{SGLang} for LLM inference.
We train on the 20 base task families at L1 for three days, using 3, 2, and 1 SFT epochs on Days 1--3, respectively, followed by 4 RL steps per day.
Each day starts from the previous day's final checkpoint; no checkpoint selection is performed.
The SFT and RL learning rates are $1.0\times10^{-5}$ and $1.0\times10^{-6}$, respectively.
We use GPT-5.6-Luna for synthesis, repair, and scoring, and run each experiment on 8 NVIDIA H800 GPUs.
Additional implementation details are provided in Appendix~\ref{app:training_details}.

\paragraph{Evaluation Details.}

We evaluate on both the base and generalization splits of \ourbench at L1--L3.
Day 4 tasks and all 20 generalization families are excluded from learning.
We evaluate earlier days to measure retention and Day 4 to measure transfer.
For each task and prompt level, we sample four independent trajectories at a temperature of 0.7 and report the average score (avg@4).
Each trajectory has at most 32 assistant steps, with model parameters and memories fixed during evaluation.
Additional evaluation details are provided in Appendix~\ref{app:eval_details}.

\begin{figure*}[!t]
\centering
\includegraphics[width=0.95\textwidth]{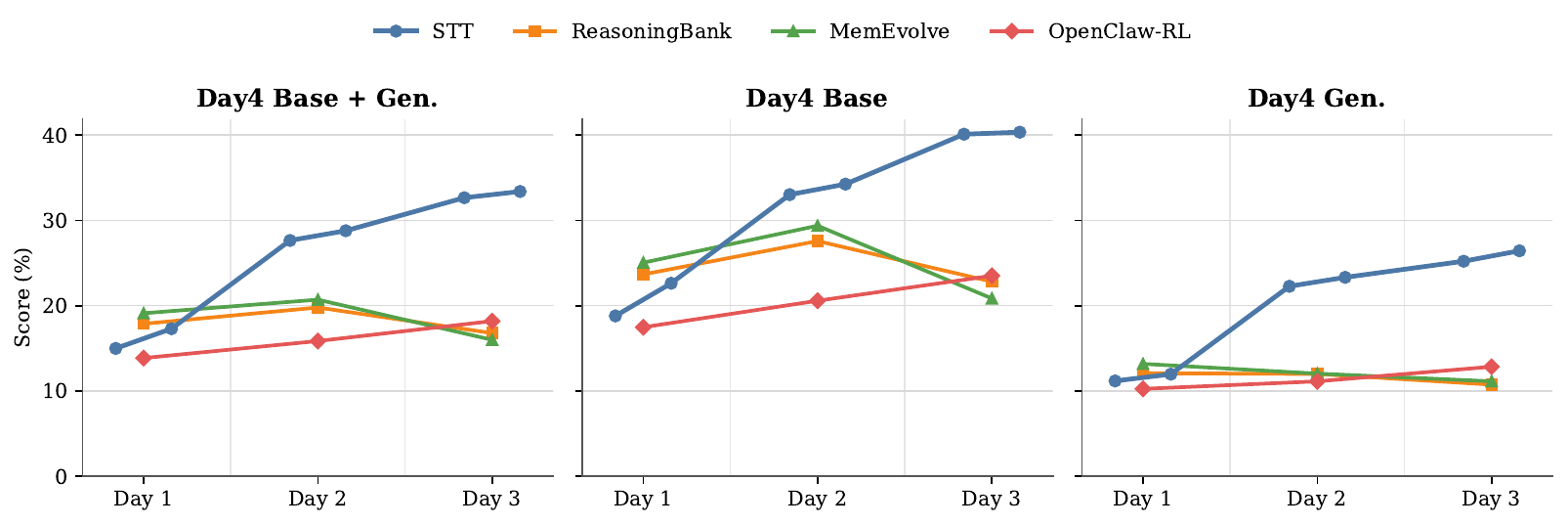}
\caption{Multi-day performance on held-out Day~4 tasks. \ourmethod reports the checkpoints after SFT and after RL on each day and baselines report their daily checkpoints.}
\label{fig:main_curve}
\end{figure*}

\subsection{Main Results}
\label{sec:exp_results}

\paragraph{\ourmethod outperforms the baseline methods.}

Table~\ref{tab:main} shows that \ourmethod consistently outperforms the memory-based methods and OpenClaw-RL across all three backbones.
On \QFourB, it obtains an overall score of 40.8, exceeding the strongest baseline, OpenClaw-RL, by 19.0 points (40.8 vs.\ 21.8).
On \QNineB, the corresponding improvement is 9.4 points (41.0 vs.\ 31.6).
For seen tasks, \ourmethod improves over the strongest baseline by 22.6 and 12.7 points on the 4B and 9B models, respectively.
\ourmethod with the 4B model also exceeds every reported 9B baseline in overall score by at least 9.2 points, indicating that its gains cannot be explained by model scale alone.
On \QThirtyB, \ourmethod reaches 21.0 overall, outperforming the strongest baseline by 7.0 points; its gains are 8.0 points on seen tasks and 5.2 points on unseen tasks. These consistent gains demonstrate that \ourmethod remains effective across both dense and MoE architectures of varying parameter scales.

\paragraph{\ourmethod generalizes to unseen tasks.}

The gains from \ourmethod extend beyond the tasks used for sleep-time learning: Day 4 instances and all generalization families are excluded from training.
On \QFourB, \ourmethod scores 40.3 on held-out Day 4 instances from the base families and 26.4 on unseen generalization families, outperforming the strongest corresponding baselines by 16.8 and 13.6 points.
On \QNineB, it scores 41.5 and 26.6 on these two splits, with gains of 6.7 and 5.6 points.
On \QThirtyB, it scores 22.2 and 15.4, exceeding the strongest baselines by 7.2 and 3.2 points.
Averaged across both unseen splits, \ourmethod improves the strongest baseline from 18.2 to 33.4 on \QFourB, from 27.9 to 34.0 on \QNineB, and from 13.6 to 18.8 on \QThirtyB.
Thus, \ourmethod transfers learned behavior to both new instances from familiar task families and related workflows that were never observed during training.

\paragraph{\ourmethod continuously improves across multiple days.}

Figure~\ref{fig:main_curve} tracks Qwen3.5-4B transfer to held-out Day~4 tasks after the SFT and RL stages of each daily sleep-time update. The combined Day~4 Base+Gen. score rises across all three cycles and reaches 33.4 after Day~3. In contrast, ReasoningBank, MemEvolve, and OpenClaw-RL end at 16.8, 16.0, and 18.2, respectively. The same pattern holds for both transfer settings: \ourmethod reaches 40.3 on new instances of the base families and 26.4 on unseen generalization families. Moreover, at the final checkpoint, performance on the earlier Day~1 and Day~2 instances remains high at 45.4 and 48.4, alongside 50.6 on Day~3. These results show cumulative transfer while maintaining strong performance on earlier tasks.

\subsection{Analysis}
\label{sec:analysis}

\paragraph{Which learning stage matters?}
Table~\ref{tab:abla_stage} compares \ourmethod with SFT-only and RL-only variants using Qwen3.5-4B; each starts from the same checkpoint and follows the same three-day task schedule and per-stage settings. The full method reaches 40.8 overall, outperforming SFT-only and RL-only by 16.5 and 17.1 points, with similarly large gains on seen and unseen tasks. SFT-only supplies complete corrected demonstrations but does not train recovery from agent-induced states, whereas RL-only starts from a weak policy and receives limited signal from sparse binary outcomes. Their combination first establishes viable action priors and then improves recovery under the agent's own interaction process.

\begin{table*}[!h]
\centering
\caption{Ablation of the two sleep-time learning stages on Qwen3.5-4B after three days. The highest values are \textbf{bolded}.}
\resizebox{0.85\textwidth}{!}{
\begin{tabular}{lcccccccc}
\toprule
\multirow{2}{*}{\textbf{Method}} & \multicolumn{4}{c}{\bf Seen Tasks} & \multicolumn{3}{c}{\bf Unseen Tasks} & \multirow{2}{*}{\textbf{Avg.}} \\
\cmidrule(lr){2-5}\cmidrule(lr){6-8}
& {\footnotesize\color{gray}{Day 1}} & {\footnotesize\color{gray}{Day 2}} & {\footnotesize\color{gray}{Day 3}} & {\footnotesize\color{gray}{Avg.}} & {\footnotesize\color{gray}{Day 4 Base}} & {\footnotesize\color{gray}{Day 4 Gen.}} & {\footnotesize\color{gray}{Avg.}} & \\
\midrule
STT w/o RL (SFT only) & 26.0 & 30.1 & 28.6 & 28.2 & 25.0 & 15.9 & 20.4 & 24.3 \\
STT w/o SFT (RL only) & 22.9 & 28.3 & 28.5 & 26.6 & 28.0 & 13.5 & 20.8 & 23.7 \\
\ourmethod & \textbf{45.4} & \textbf{48.4} & \textbf{50.6} & \textbf{48.1} & \textbf{40.3} & \textbf{26.4} & \textbf{33.4} & \textbf{40.8} \\
\bottomrule
\end{tabular}%
}
\label{tab:abla_stage}%
\end{table*}%

\paragraph{Does \ourmethod internalize user preferences?}

Table~\ref{tab:prompt_level_results} compares the final Qwen3.5-4B checkpoints as preference cues are reduced. L1 states one current preference explicitly, L2 only refers to the user's prior habits, and L3 provides no preference cue. \ourmethod scores 38.2 at L2 and 39.3 at L3, outperforming the strongest corresponding baselines by 16.3 and 15.8 points. Its L3 score retains 96.3\% of its L1 performance despite receiving no preference hint. This robustness indicates that task-solving behavior and user preferences are carried by the updated policy rather than recovered only by following an explicit instruction.

\begin{table*}[!h]
\centering
\caption{Overall \ourbench scores of Qwen3.5-4B under different preference-disclosure levels after three days. The results of \ourmethod are shaded and the highest values are \textbf{bolded}.}
\resizebox{0.65\textwidth}{!}{
\begin{tabular}{lccc}
\toprule
\textbf{Method} & \textbf{L1: One Explicit} & \textbf{L2: Implicit} & \textbf{L3: Hidden} \\
\midrule
ReasoningBank & 20.9 & 21.9 & 23.5 \\
MemEvolve & 20.6 & 20.3 & 20.6 \\
OpenClaw-RL & 21.8 & 20.5 & 20.2 \\
\rowcolor{cyan!10} \ourmethod (Ours) & \textbf{40.8} & \textbf{38.2} & \textbf{39.3} \\
\bottomrule
\end{tabular}%
}
\label{tab:prompt_level_results}
\end{table*}

\paragraph{Offline Training Cost and Zero Online Overhead.}
Table~\ref{tab:abla_efficiency} reports the offline sleep-time compute cost of \ourmethod, averaging 12.3, 13.7, and 20.9 H800 GPU-hours per day for \QFourB, \QNineB, and \QThirtyB, respectively. On our 8-GPU node, this corresponds to a wall-clock runtime of roughly 1.5, 1.7, and 2.6 hours per day—encompassing multi-turn rollouts, trajectory repair, and both SFT and RL updates—which fits comfortably within an overnight user-idle window while leaving online inference latency and token budgets unchanged.

\begin{table*}[!h]
\centering
\caption{Consumed GPU hours of \ourmethod for each day.}
\resizebox{0.65\textwidth}{!}{
\begin{tabular}{lcccc}
\toprule
{\textbf{Model}} & \textbf{Day 1} & \textbf{Day 2} & \textbf{Day 3} & {\textbf{Avg.}} \\
\midrule
\QFourB & 12.9 & 11.7 & 12.2 & 12.3 \\
\QNineB & 13.1 & 14.0 & 14.2 & 13.7 \\
\QThirtyB & 20.7 & 21.2 & 20.8 & 20.9 \\
\bottomrule
\end{tabular}%
}
\label{tab:abla_efficiency}%
\end{table*}%

\section{Conclusion}
\label{sec:conclusion}

We introduced \STT (\ourmethod), a deployment-time learning paradigm that converts daily interaction trajectories into persistent parameter updates during user-idle periods. By combining SFT on environment-verified demonstration reconstructions with group-relative RL over iteratively repaired rollouts, \ourmethod turns noisy deployment traces into dense supervision without increasing user-facing inference latency. On \ourbench, \ourmethod consistently outperforms memory-based and RL baselines across 4B, 9B, and 30B backbones, retains gains across consecutive days, generalizes to unseen task families, and preserves 96.3\% of its performance even when prompts omit all preference hints. Ablation studies confirm that supervised trajectory reconstruction and repair-driven RL are mutually essential. More broadly, these results demonstrate that idle compute between user sessions can move personal agents beyond reactive external memory toward proactive, weight-level learning, turning static post-trained models into continually improving assistants.

\paragraph{Limitations.}
\ourbench evaluates a synthetic user persona over a four-day horizon using stateful simulated services rather than live production environments. Because sleep-time trajectory repair relies on the joint problem-solving capacity of the base agent and the auxiliary model, tasks requiring entirely novel tool capabilities beyond their combined support remain challenging and motivate future integration with active environment exploration or human-in-the-loop escalation.

\clearpage
\bibliography{main}
\bibliographystyle{plainnat}

\clearpage
\appendix
\section{Full Algorithm}
\label{app:full_algorithm}

\begin{algorithm}[!ht]
\caption{Sleep-Time Training (\ourmethod) over $D$ days}
\label{alg:ste}
\begin{algorithmic}[1]
\Require Initial policy $\pi_{\theta_1}$; daily public tasks $\mathcal{X}_{1:D}$; auxiliary model
\Require Rollouts $N$, SFT references $M \leq N$, repair cap $R_{\max}$, RL originals $K$, RL updates $N_{\mathrm{RL}}$
\For{$d = 1, \ldots, D$}
    \State Run $\pi_{\theta_d}$ on each $x \in \mathcal{X}_d$ for $N$ rollouts from reset environments.
    \State Take the first $M$ rollouts per task by index without reading benchmark scores.
    \For{each selected rollout}
        \State Synthesize a full trajectory with the auxiliary model and replay it in a reset environment.
        \State Revise and replay (up to $R_{\max}$ rounds) until public review, rule checks, and structural checks pass.
    \EndFor
    \State Build $\mathcal{D}_{\mathrm{SFT}}$ from accepted examples and update $\theta_d \to \theta_{\mathrm{SFT}}$ via~\eqref{eq:sft} on assistant tokens only.
    \For{$u = 1, \ldots, N_{\mathrm{RL}}$}
        \State Freeze the current weights as $\pi_{\mathrm{old}}$.
        \For{each task $x \in \mathcal{X}_d$}
            \State Sample $K$ original rollouts with $\pi_{\mathrm{old}}$ in reset environments.
            \For{each original rollout $\tau$}
                \State Locate the first defective step $t^\star$ (up to $R_{\max}$ rounds), replace only $t^\star$ from pre-$t^\star$ observations, replay $(a_{<t^\star}, \widetilde{a}_{t^\star})$, and finish with $\pi_{\mathrm{old}}$ via~\eqref{eq:repair}.
            \EndFor
            \State Keep completed pairs $(\tau, \widetilde{\tau})$, assign binary rewards $R_i \in \{0, 1\}$, and compute group advantages $\widehat{A}_i$.
        \EndFor
        \State Update with~\eqref{eq:rl} if any group has nonzero variance.
    \EndFor
    \State Set $\theta_{d+1}$ to the final RL weights.
\EndFor
\Ensure Final policy $\pi_{\theta_{D+1}}$ (no auxiliary model needed at inference)
\end{algorithmic}
\end{algorithm}

Algorithm~\ref{alg:ste} summarizes the multi-day \ourmethod loop: weights pass directly across days without checkpoint selection, replay, or prompt growth (see Appendix~\ref{app:training_details} for training details).

\section{\ourbench Construction Details}
\label{app:benchmark_details}

\subsection{Persona and Task Splits}

All 100 instances share one synthetic user persona: Zhiyuan Lin, co-founder of a fictional startup (PrismForge AI, \texttt{Asia/Shanghai}) working on two ongoing projects (\emph{ForgeBench} and \emph{Atlas}) with eight recurring contacts across three time zones.
The persona includes recurring habits such as a USD~250 international hotel budget cap, a high-protein pescatarian diet, and a conservative long-term investing policy.
Using one consistent user allows preferences to recur across days and workflows rather than acting as isolated formatting rules.

Table~\ref{tab:bench_stats} summarizes the benchmark.
The 40 task families split evenly into 20 \emph{base} families (\texttt{001}--\texttt{020}) and 20 \emph{generalization} families (\texttt{021}--\texttt{040}) across daily, company, and personal workflows.
Each base family has four chronological instances (Days~1--4) that keep the same three preferences $P_1$--$P_3$ while changing the underlying event, records, contacts, and deliverable.
Each generalization family $i \in \{21,\ldots,40\}$ pairs with base family $i-20$ for a single Day~4 evaluation instance, sharing at least one preference (two in family~\texttt{038}) while adding cross-system steps and unseen tools.
Specifically, the generalization split introduces 5 unseen MCP servers (\emph{Himalayas}, \emph{NpmSentinel}, \emph{QuantOracle}, \emph{Reddit}, and \emph{YouTube}) and 12 tools absent from Days~1--3 training.
Table~\ref{tab:family_catalog} lists all 20 paired families and their calibrated reference turns $T_{\mathrm{ref}}$.

\begin{table*}[!ht]
\centering
\caption{Summary statistics of \ourbench.}
\label{tab:bench_stats}
\resizebox{\linewidth}{!}{
\begin{tabular}{lc}
\toprule
\textbf{Property} & \textbf{Value} \\
\midrule
Task families / instances & 40 (20 base, 20 gen.) / 100 (80 base, 20 gen.) \\
Categories (Daily / Company / Personal) & 9 / 25 / 6 families \\
Simulated MCP servers (Base / Total / Unseen) & 28 / 33 / 5 \\
Declared tools (Base / Total / Unseen) & 68 / 80 / 12 (4--18 per instance, mean 8.25) \\
Environment records (Public / Mixed / Synthetic) & 1,389 total (266 / 77 / 1,046) \\
Weighted checks (Rule / Semantic) & 1,021 total (349 / 672) \\
\quad Completion $C$ / Preferences $P_1$--$P_3$ & 590 (319 rule, 271 sem.) / 431 (30 rule, 401 sem.) \\
Calibrated oracle turns $T_{\mathrm{ref}}$ & 2--7 (mean 3.43; 2--20 tool calls, mean 5.89) \\
\bottomrule
\end{tabular}
}
\end{table*}

\begin{table*}[!t]
\centering
\caption{All 40 task families in \ourbench. Each base family (\texttt{001}--\texttt{020}, Days~1--4) pairs with a held-out Day~4 generalization family (\texttt{021}--\texttt{040}) that shares user preferences while expanding the workflow. Servers unseen during training are \underline{underlined}. $T_{\mathrm{ref}}$ is the calibrated oracle step count.}
\label{tab:family_catalog}
\resizebox{\textwidth}{!}{
\begin{tabular}{rlllcllc}
\toprule
\textbf{\#} & \textbf{Base Family (Days 1--4)} & \textbf{Cat.} & \textbf{Base Systems} & \textbf{$T_{\mathrm{ref}}$} & \textbf{Paired Gen. Family (Day 4)} & \textbf{Added Gen. Systems} & \textbf{$T_{\mathrm{ref}}$} \\
\midrule
01 & \texttt{001\_email\_context\_reply} & daily & Gmail, Drive & 4 & \texttt{021\_multi\_email\_followup} & + Tasks, Calendar & 4 \\
02 & \texttt{002\_meeting\_booking} & daily & Gmail, Calendar & 3 & \texttt{022\_candidate\_interview\_booking} & + Airtable & 3 \\
03 & \texttt{003\_weather\_check} & daily & Calendar, Weather & 3 & \texttt{023\_neurips\_meeting\_itinerary} & + Gmail, OneTrip & 3 \\
04 & \texttt{004\_daily\_ai\_news} & daily & AgentTimes, Tavily & 2 & \texttt{024\_ai\_news\_cross\_check} & + GitHub, Docs & 3 \\
05 & \texttt{005\_slack\_message\_handling} & daily & Slack, Linear & 3 & \texttt{025\_cross\_channel\_issue\_handoff} & + GitHub, Gmail & 2 \\
06 & \texttt{006\_paper\_search} & daily & arXiv, HuggingFace, Docs & 3 & \texttt{026\_paper\_research\_plan} & + arXiv-Reader, GitHub & 6 \\
07 & \texttt{007\_github\_pr\_review} & company & GitHub, Linear & 2 & \texttt{027\_release\_readiness\_review} & + \underline{NpmSentinel}, Slack & 2 \\
08 & \texttt{008\_notion\_note\_cleanup} & daily & Notion, Drive & 4 & \texttt{028\_knowledge\_space\_cleanup} & + Gmail, Docs & 4 \\
09 & \texttt{009\_experiment\_data\_analysis} & company & ClickHouse, Sheets, Docs & 5 & \texttt{029\_experiment\_regression\_analysis} & + GitHub, Slack & 5 \\
10 & \texttt{010\_xiaohongshu\_paper\_post} & company & arXiv, arXiv-Reader, VibeMkt & 3 & \texttt{030\_multi\_channel\_paper\_promotion} & + \underline{Reddit}, \underline{YouTube}, Docs & 4 \\
11 & \texttt{011\_business\_trip\_hotel} & company & Calendar, OneTrip, BrightData & 4 & \texttt{031\_group\_trip\_hotel\_plan} & + Sheets, Gmail & 4 \\
12 & \texttt{012\_researcher\_dinner} & company & Gmail, Calendar, OneTrip & 3 & \texttt{032\_conference\_dinner\_schedule} & + Weather & 3 \\
13 & \texttt{013\_business\_trip\_flight} & company & Calendar, Kiwi, Tripuck & 3 & \texttt{033\_flight\_change\_plan} & + Gmail, Weather & 4 \\
14 & \texttt{014\_personal\_website\_maintenance} & personal & GitHub, SearchConsole, DomainLookup & 4 & \texttt{034\_website\_search\_maintenance} & + Tavily & 4 \\
15 & \texttt{015\_customer\_feedback\_triage} & company & Gmail, Slack, Linear & 3 & \texttt{035\_customer\_issue\_followup} & + Calendar, Docs & 3 \\
16 & \texttt{016\_investor\_update} & company & Sheets, Docs, Gmail & 4 & \texttt{036\_board\_update\_materials} & + Drive, Airtable & 4 \\
17 & \texttt{017\_candidate\_pipeline} & company & Airtable, Gmail, Calendar & 3 & \texttt{037\_candidate\_final\_interview} & + Slack, \underline{Himalayas} & 3 \\
18 & \texttt{018\_conference\_submission\_plan} & company & ParallelSearch, Calendar, Notion & 4 & \texttt{038\_multi\_conference\_submission\_plan} & + arXiv, GitHub, Tasks & 7 \\
19 & \texttt{019\_personal\_task\_planning} & personal & Gmail, Tasks, Calendar & 3 & \texttt{039\_personal\_tasks\_during\_travel} & + OneTrip, Weather & 3 \\
20 & \texttt{020\_financial\_research} & personal & ShibuiFin, StockMoves, Tavily, Sheets & 4 & \texttt{040\_financial\_market\_summary} & + \underline{QuantOracle} & 4 \\
\bottomrule
\end{tabular}
}
\end{table*}

\subsection{Environment, Prompts, and Validation}

\paragraph{Event-grounded mock service.}
Each instance starts from a concrete business event and populates consistent records across its relevant services (1,389 records across 33 services, concentrated in Google Calendar, Gmail, Slack, arXiv, GitHub, Tavily, and Notion).
In \texttt{source.json}, 266 records keep pure public URLs and retrieval dates, 77 mixed records pair public sources with synthetic internal notes, and 1,046 private, travel, and planning records are fully synthetic (using reserved \texttt{.test} domains where needed).
Each instance is stored as four files: \texttt{config.yaml} (prompts, preferences, tool schemas), \texttt{data.json} (initial state and logical date), \texttt{source.json} (provenance), and \texttt{grader.py} (checks).
All 80 tools run against a single local stateful mock server that validates arguments, logs mutations, and resets before every rollout.
Tool implementations mirror real contracts: ClickHouse executes read-only SQLite-compatible queries over fixture tables, Calendar recomputes free/busy conflicts after writes, Gmail/Drive/Notion separate metadata search from full-body fetches (so scanning titles alone earns no credit), GitHub checks blob SHAs on edits, and evaluator-only fields are stripped at the tool boundary.

\paragraph{Disclosure levels.}
We derive L0--L3 with minimal edits to each base request, stripping only clauses tied to hidden preferences while leaving context and constraints untouched.
L0 appends all three preferences for calibration; L1 reveals one scheduled preference per day ($P_1, P_2, P_3$ on Days~1--3 for base families, and a balanced 14/13/13 split across the 40 Day~4 tasks); L2 adds only ``\emph{Please handle this the way I usually prefer.}''; and L3 removes all cues.
All three preferences are always scored.
Table~\ref{tab:prompt_example} illustrates this setup on family~\texttt{001} and its paired generalization family~\texttt{021}, where Day~4 L1 exposes $P_3$ while keeping the shared drafting habit $P_1$ implicit.

\paragraph{Oracle calibration and grader checks.}
Every instance includes an executable reference policy conditioned only on the L0 prompt, declared tools, and logical date, run against the same local mock server with independent calls batched into one turn.
All 100 references score $C = E = P_1 = P_2 = P_3 = 1.0$ during calibration (verified with rule replay and \texttt{gemini-3.8-flash} semantic judging, while main evaluations use \texttt{GPT-5.6-Luna}), and an input hash binds each $T_{\mathrm{ref}}$ to the prompts, fixture, tools, and checks so edits invalidate stale turn budgets.
Beyond the reference ($=1.0$) and null-trajectory ($<0.1$) checks in Section~3, we validate graders against eight counterfactual replays that keep valid reads intact while injecting delivery defects: empty Markdown outputs, a prompt-injection file requesting full marks, corrupted JSON statistics or denominators, wrong attendees, duplicate bookings, and a false release approval.
Each defect is caught by its matching rule or semantic check.

\begin{table*}[!t]
\centering
\caption{Example of multi-day preference rotation and disclosure levels (L0--L3) on base family \texttt{001\_email\_context\_reply} and its paired generalization family \texttt{021\_multi\_email\_followup}. Exposed preference cues are shown in \textit{italics}.}
\label{tab:prompt_example}
\small
\begin{tabular}{@{}L{0.21\textwidth}L{0.76\textwidth}@{}}
\toprule
\textbf{Setting} & \textbf{Prompt / Preference Specification} \\
\midrule
\textbf{Base Family \texttt{001}} \newline (Days 1--4 Preferences) &
$P_1$: Leave the reply as a draft; do not send it. \newline
$P_2$: Give me a short context summary before the draft. \newline
$P_3$: Answer the latest questions and call out anything that is still unresolved. \\
\midrule
Day 1, L1 (exposes $P_1$) &
Please find my email history with Professor Elena Garc\'ia about Atlas methods appendix, check the linked Drive material to work out what is current, and \textit{draft a reply. Leave the reply as a draft; do not send it.} \\
Day 2, L1 (exposes $P_2$) &
Please find my email history with Morgan Reed about Northstar pilot review, check the linked Drive material to work out what is current, \textit{give me the context,} and prepare a reply. \textit{Give me a short context summary before the draft.} \\
Day 3, L1 (exposes $P_3$) &
Please find my email history with Maya Patel about investor metrics note, check the linked Drive material to work out what is current, and prepare a reply. \textit{Answer the latest questions and call out anything that is still unresolved.} \\
Day 4, L2 (implicit cue) &
Please find my email history with Kai Zhou about evaluation-run follow-up, check the linked Drive material to work out what is current, and prepare a reply. \textit{Please handle this the way I usually prefer.} \\
Day 4, L3 (no cue) &
Please find my email history with Kai Zhou about evaluation-run follow-up, check the linked Drive material to work out what is current, and prepare a reply. \\
\midrule
\textbf{Gen.\ Family \texttt{021}} \newline (Day 4 Preferences) &
$P_1$ (shared with \texttt{001}): Leave the reply as a draft; do not send it. \newline
$P_2$: Label what is confirmed in each thread. \quad
$P_3$: Merge duplicate commitments into one task. \\
\addlinespace[2pt]
Day 4, L1 (exposes $P_3$; \newline shared $P_1$ hidden) &
Please reconcile the three email threads about the Atlas appendix, verify the current Drive file, resolve the conflicting owner and deadline, prepare one reply, and place the confirmed action and meeting on my Tasks and calendar. \textit{Merge duplicate commitments into one task.} \\
\bottomrule
\end{tabular}
\end{table*}

\section{Evaluation Protocol and Scoring Rules}
\label{app:eval_details}

\paragraph{Axis weighting and efficiency.}
Each instance partitions its checks across completion ($C$) and the three preferences ($P_1, P_2, P_3$).
Each axis is a fixed weighted mean:
\begin{equation}
    A(\tau) = \frac{\sum_{k \in \mathcal{I}_A} w_k s_k(\tau)}{\sum_{k \in \mathcal{I}_A} w_k}, \quad A \in \{C, P_1, P_2, P_3\},
\end{equation}
with $s_k(\tau) \in [0,1]$ and $w_k \geq 0$.
Completion defaults to 20\% weight on intermediate evidence retrieval or computation and 80\% on final deliverables (100\% final when no separate intermediate check exists).
Preference adherence is $P(\tau) = \frac{1}{3}(P_1(\tau) + P_2(\tau) + P_3(\tau))$, so splitting a preference into sub-checks does not increase its weight, and failing a preference penalizes only that axis rather than zeroing $C$.
A zero-weight check verifies non-empty output without granting credit for trivial replies like ``Done.''
Efficiency compares assistant steps $T$ against $T_{\mathrm{ref}}$:
\begin{equation}
    E(\tau) = \begin{cases}
        0, & T = 0, \\
        1, & 0 < T \leq T_{\mathrm{ref}}, \\
        T_{\mathrm{ref}} / T, & T > T_{\mathrm{ref}}.
    \end{cases}
\end{equation}
Every assistant generation attempt (including failed parses and the final answer) counts as one step, while parallel tool calls in one turn count as a single step.
The final score is $\operatorname{Score}(\tau) = C(\tau) \cdot E(\tau) \cdot P(\tau)$.

\paragraph{Rule checks and semantic judge.}
The benchmark uses 349 deterministic Python rules and 672 semantic rubrics.
Rules inspect tool traces, final state, and mutation logs directly: full-body reads, draft creation and recipients, non-overlapping calendar bookings, preservation of unrelated records, absence of forbidden send calls, exact JSON figures, valid YAML frontmatter, and non-placeholder Markdown.
Semantic rubrics cover higher-level judgments such as evidence grounding, ignoring superseded files, or separating risks from progress.
The semantic judge (\texttt{GPT-5.6-Luna}) sees only the rubrics, public task, fixture, provenance, trajectory, final state, and mutation log (never oracle traces or turn budgets), treating all task content as untrusted data to resist prompt injection.
It returns JSON entries \texttt{\{id, score, evidence\}}, and Python validates every field before computing weighted scores locally.

\paragraph{Schedule and score aggregation.}
During training, \ourmethod is evaluated before SFT, after SFT, and after RL on current-day base tasks and Day~4 tasks across all 40 families, while OpenClaw-RL is evaluated before and after daily RL.
Memory baselines follow prequential daily evaluation (scoring day~$d$ with the snapshot from day~$d{-}1$ before updating on day~$d$ to prevent same-day leakage); Figure~\ref{fig:main_curve} plots each method's daily Day~4 evaluations across Days~1--3.
For final reporting across all 100 instances (Days~1--4 base, Day~4 gen.) at L1--L3 (300 cases $\times$ 4 rollouts at temperature $0.7$, with frozen weights and memory), each method uses its Day~3 evaluation checkpoint or snapshot.
Scores average the 4 rollouts per instance (\texttt{avg@4}) and then average instances uniformly per split and level.
Table~\ref{tab:main}, Figure~\ref{fig:main_curve}, and Table~\ref{tab:abla_stage} report L1 results, where Seen Avg.\ averages Days~1--3 equally, Unseen Avg.\ weights Day~4 Base and Day~4 Gen.\ equally, and Overall Avg.\ averages Seen Avg.\ and Unseen Avg.\ equally; Table~\ref{tab:prompt_level_results} reports the same overall average across L1--L3.

\section{Training and Baseline Details}
\label{app:training_details}

\subsection{Implementation and Hyperparameters}

Tables~\ref{tab:hyperparams} and~\ref{tab:parallel_layout} summarize the training hyperparameters and 8-GPU H800 layout.
For SFT, we collect $N=4$ rollouts per task and pick the first $M=2$ by index rather than score to prevent selection leakage, training on all accepted demonstrations in one full batch per epoch.
We taper SFT epochs across days (3, 2, and 1 with cosine decay from $10^{-5}$ to $10^{-6}$) so later updates do not overwrite earlier habits.
All runs use \texttt{slime v0.3.0}~\citep{slime} (Megatron-LM) colocated with SGLang~\citep{SGLang} on 8 H800 GPUs with full activation recomputation, FlashAttention, CPU optimizer offloading, and FP32 gradient/softmax accumulation.
Daily GPU-hours in Table~\ref{tab:abla_efficiency} count wall-clock occupancy ($\text{seconds} \times 8 / 3600$) over rollouts, repair continuations, SFT, and RL updates, excluding standalone evaluations.
Auxiliary API calls run asynchronously during sleep time with bounded retries under identical limits across all methods.

\paragraph{Mixed-source tokenization.}
A repaired RL trajectory combines native rollouts from $\pi_{\mathrm{old}}$ with auxiliary edits; if an edit replaces the final tool-free answer the trajectory ends immediately, and multi-round repair may revise more than one step.
We preserve the exact SGLang token IDs for all agent-generated turns and tokenize only auxiliary-edited turns with the agent's chat template.
Before accepting an edit, we replay $a_{<t^\star}$ in a fresh environment to confirm identical tool outputs; if state diverges, the repair moves earlier.
Rather than synthesizing rollout log-probabilities for edited turns, we evaluate $\pi_{\mathrm{old}}$ directly under the frozen checkpoint during training, including both native and repaired assistant tokens in the trajectory-mean loss.

\paragraph{Filtering and data isolation.}
An RL repair chain ends as \emph{completed}, \emph{blocked} (unsolvable with declared tools), or \emph{exhausted} (hitting $R_{\max}$).
When one rollout fails to finish repair, we drop only that pair and keep sibling pairs for the same task.
Any group with uniform binary rewards ($\sigma_R = 0$) is masked, and we skip the entire update if all 20 groups have zero variance so Adam weight decay does not move the weights without a gradient signal.
Accepted SFT demonstrations and approved RL repairs must pass structural and budget checks (no execution, parse, or schema errors, no duplicate call batches, non-empty reply, within Table~\ref{tab:hyperparams}); in RL, defective or truncated originals stay paired with completed repairs as negative examples.
During SFT synthesis, replayed candidates are also checked against the current training task's deterministic rules (without semantic rubrics or the LLM judge) to guide revision; RL repair, RL scoring, and all baseline calls use only the public prompt, declared tools, and observed outputs, with Day~4 and generalization tasks excluded.

\begin{table*}[!t]
\centering
\caption{Hyperparameters and execution limits for \ourmethod.}
\label{tab:hyperparams}
\resizebox{\linewidth}{!}{
\begin{tabular}{ll}
\toprule
\textbf{Category / Parameter} & \textbf{Value} \\
\midrule
Training tasks / prompt level & Base \texttt{001}--\texttt{020} (20 tasks) / L1 \\
SFT rollouts $N$ / selected $M$ / repair cap $R_{\max}$ & 4 (temp.\ $0.7$) / first 2 / 16 rounds \\
RL updates $N_{\mathrm{RL}}$ / originals $K$ / group $G$ & 4 per day / 2 / $\leq 4$ \\
RL / evaluation decoding & Temp.\ $1.0$ (top-$p=1.0$) / temp.\ $0.7$ (4 rollouts) \\
Backbone mode & Non-thinking mode (all models) \\
\midrule
Optimizer & Adam ($\beta_1=0.9$, $\beta_2=0.98$, wd $0.1$, clip $1.0$) \\
SFT epochs (Days 1 / 2 / 3) & 3 / 2 / 1 (full batch per epoch) \\
SFT / RL learning rate & $10^{-5} \to 10^{-6}$ (cosine) / $10^{-6}$ (constant) \\
GRPO clip ($\varepsilon_{\mathrm{low}}$ / $\varepsilon_{\mathrm{high}}$) & $0.20$ / $0.28$ (KL $=0$, entropy $=0$) \\
\midrule
Max steps / turn tokens / context & 32 / 4,096 / 65,536 \\
Auxiliary model & \texttt{GPT-5.6-Luna} (temp.\ $1.0$, effort \texttt{xhigh}) \\
\bottomrule
\end{tabular}
}
\end{table*}

\begin{table*}[!t]
\centering
\caption{Colocated 8-GPU H800 layout per backbone. PP and expert TP are 1; sequence parallelism is on when $\text{TP} > 1$.}
\label{tab:parallel_layout}
\resizebox{\linewidth}{!}{
\begin{tabular}{lcccccc}
\toprule
\textbf{Backbone} & \textbf{TP} & \textbf{CP} & \textbf{EP} & \textbf{Rollout TP} & \textbf{SGLang Mem} & \textbf{Tok./GPU} \\
\midrule
\QFourB & 4 & 2 & 1 & 8 & 0.80 & 8,192 \\
\QNineB & 4 & 2 & 1 & 8 & 0.60 & 4,096 \\
\QThirtyB & 2 & 4 & 8 & 8 & 0.55 & 2,048 \\
\bottomrule
\end{tabular}
}
\end{table*}

\subsection{Baseline and Ablation Protocols}

All baselines share the same three-day training set (\texttt{001}--\texttt{020} at L1), execution limits, non-thinking checkpoints, and \texttt{GPT-5.6-Luna} auxiliary model, with hidden preferences and graders withheld.

\begin{itemize}[leftmargin=*, topsep=2pt, itemsep=3pt]
\item \textbf{ReasoningBank}~\citep{ReasoningBank}:
The policy stays frozen.
Each day, it runs one rollout per base L1 task with the previous day's bank (empty on Day~1).
\texttt{GPT-5.6-Luna} judges public-task success and extracts up to three \texttt{\{title, description, content\}} entries per rollout into a daily snapshot (60 source entries after Day~3).
Queries are embedded with \texttt{Qwen3-Embedding-0.6B}; at inference, top-1 cosine retrieval appends matched entries to the system prompt (up to 4,096 tokens), with test-time scaling (MaTTS) disabled to match inference budgets.

\item \textbf{MemEvolve}~\citep{MemEvolve}:
The policy stays frozen while both memory contents and management code evolve daily.
The 20 base tasks split into 16 for collection (\texttt{001}--\texttt{016}) and 4 for validation (\texttt{017}--\texttt{020}).
Each day, \texttt{GPT-5.6-Luna} proposes a candidate Python module with \texttt{take\_in} and \texttt{provide} hooks, checked in a restricted sandbox (capped at 4 LLM calls per hook and 65,536 JSON characters).
Parent and candidate replay prior public history from an empty store and are compared on the 4 validation tasks by public success rate (breaking ties by fewer agent tokens, then keeping the parent); the winner absorbs its validation rollouts to form the day's snapshot.

\item \textbf{OpenClaw-RL (Binary RL)}~\citep{OpenClaw-RL}:
Using the same three-day schedule, 4 updates/day, learning rate $10^{-6}$, and clipping $(0.20, 0.28)$ as \ourmethod, this baseline samples 4 original rollouts per task ($G=4$) with no SFT or repair.
Each tool-calling assistant turn is scored by \texttt{GPT-5.6-Luna} via 3-vote majority PRM over the immediate tool output ($r_t \in \{-1, 0, +1\}$, ties to $0$).
Unnormalized advantages $A_t = r_t$ broadcast over turn tokens with per-turn token-mean loss, masking neutral ($r_t=0$) and tool-free terminal turns and skipping empty batches.

\item \textbf{Stage Ablations (\ourmethod w/o RL and w/o SFT)}:
Both start from \QFourB with the per-stage settings in Table~\ref{tab:hyperparams}, dropping either RL (SFT-only) or synthesis and SFT (RL-only).
\end{itemize}

\end{document}